\pdfoutput=1

\documentclass[11pt]{article}

\usepackage{acl}

\usepackage{times}
\usepackage{latexsym}

\usepackage[T1]{fontenc}

\usepackage[utf8]{inputenc}

\usepackage{microtype}

\usepackage{graphicx}

\title{\emph{PAI-Diffusion}: Constructing and Serving a Family of Open Chinese Diffusion Models for Text-to-image Synthesis on the Cloud}

\author{Chengyu Wang\textsuperscript{\rm 1},
    Zhongjie Duan\textsuperscript{\rm 1,2},
    Bingyan Liu\textsuperscript{\rm 1,3},
    Xinyi Zou\textsuperscript{\rm 1},
    Cen Chen\textsuperscript{\rm 2},
    Kui Jia\textsuperscript{\rm 3},
    Jun Huang\textsuperscript{\rm 1\thanks{\ \ Corresponding author.}} \\
  \textsuperscript{\rm 1} Alibaba Group, Hangzhou, China\\
  \textsuperscript{\rm 2} East China Normal University, Shanghai, China\\
  \textsuperscript{\rm 3} South China University of Technology, Guangzhou, China\\
  \texttt{\{chengyu.wcy,zouxinyi.zxy,huangjun.hj\}@alibaba-inc.com}\\
  \texttt{eeliubingyan@mail.scut.edu.cn,zjduan@stu.ecnu.edu.cn}\\
  \texttt{cenchen@dase.ecnu.edu.cn,kuijia@gmail.com}\\}

\begin{document}
\maketitle
\begin{abstract}
Text-to-image synthesis for the Chinese language poses unique challenges due to its large vocabulary size, and intricate character relationships. While existing diffusion models have shown promise in generating images from textual descriptions, they often neglect domain-specific contexts and lack robustness in handling the Chinese language. This paper introduces~\emph{PAI-Diffusion}, a comprehensive framework that addresses these limitations.~\emph{PAI-Diffusion} incorporates both general and domain-specific Chinese diffusion models, enabling the generation of contextually relevant images. It explores the potential of using LoRA and ControlNet for fine-grained image style transfer and image editing, empowering users with enhanced control over image generation. Moreover,~\emph{PAI-Diffusion} seamlessly integrates with Alibaba Cloud's Machine Learning Platform for AI, providing accessible and scalable solutions. All the Chinese diffusion model checkpoints, LoRAs, and ControlNets, including domain-specific ones, are publicly available. A user-friendly Chinese WebUI and the~\emph{diffusers-api} elastic inference toolkit, also open-sourced, further facilitate the easy deployment of~\emph{PAI-Diffusion} models in various environments, making it a valuable resource for Chinese text-to-image synthesis.
\end{abstract}

\section{Introduction}

Recently, diffusion models~\cite{DBLP:conf/cvpr/RombachBLEO22,DBLP:conf/nips/SahariaCSLWDGLA22} have emerged to address the challenges of generating realistic and high-quality images from textual descriptions. 
This research area has obtained widespread attention, driven by the increasing demand for automated image synthesis in various applications, such as art design, virtual reality, etc~\cite{DBLP:journals/corr/abs-2302-06826,DBLP:journals/corr/abs-2302-02284}.

In the community, Stable Diffusion\footnote{\url{https://stability.ai/stablediffusion}} has gained significant popularity due to its ability to generate high-quality images that align well with textual descriptions. However, when it comes to handling the Chinese language, similar models encounter certain challenges that hinder its performance. From the linguistic aspect, Chinese is a morphologically rich language with a large vocabulary size and complex inter-dependencies between characters. Furthermore, Chinese characters often have multiple meanings and can be combined to form compound words, making it challenging to establish accurate and consistent mappings between textual descriptions and visual representations~\cite{DBLP:conf/emnlp/Liu0ZLQ00X22}.

Previously, in the literature, several works have been proposed to make specialized adaptations of diffusion models to improve the performance of text-to-image synthesis for Chinese~\cite{DBLP:journals/corr/abs-2209-02970,DBLP:conf/acl/ChenLZYW23,DBLP:journals/corr/abs-2305-11540}. Yet, there are still some notable drawbacks that need to be addressed.
i) Many existing models focus on generating images based on generic textual descriptions, neglecting the ability to generate images in specific domains or contexts. ii) For the Chinese language, the potential of using LoRA~\cite{DBLP:conf/iclr/HuSWALWWC22} and ControlNet~\cite{DBLP:journals/corr/abs-2302-05543} for fine-grained image style transfer and image editing have not been fully explored.
iii) The lack of support for cloud-based product integration is another important drawback. Given the increasing popularity of cloud-based services and the demand for scalable and accessible text-to-image solutions, it is crucial to develop models and solutions that can be easily integrated into cloud platforms or deployed as cloud-based services.

In this work, we formally present~\emph{PAI-Diffusion}, which consists of a family of open Chinese diffusion models, together with user-friendly toolkits to serve these models on the cloud. Major features of~\emph{PAI-Diffusion} include the following:
\begin{itemize}
    \item \emph{PAI-Diffusion} incorporates both general and domain-specific Chinese diffusion models, specifically allowing for the generation of images that are tailored to specific contexts or domains (such as Chinese cuisine, poetry and paintings). This enables users to create visually compelling and relevant images for various domain-specific applications.

    \item \emph{PAI-Diffusion} explores the potential of using LoRA and ControlNet for fine-grained style transfer and image editing, with a variety of corresponding Chinese models released. This empowers users with greater control over the generated images, enabling them to manipulate fine-grained semantic attributes and modify visual features based on their preferences.

    \item \emph{PAI-Diffusion} extends our previous work~\cite{DBLP:conf/acl/LiuLD0WZJJC023} and ensures seamless integration with our Machine Learning Platform for AI (PAI) of Alibaba Cloud\footnote{\url{https://www.alibabacloud.com/product/machine-learning}}, providing users with accessible and scalable solutions. By integrating with cloud products,~\emph{PAI-Diffusion} enables users to harness the power of cloud computing and enjoy the benefits of user-friendly and resource-efficient systems. 
    For designers, our Chinese WebUI toolkit largely extends the Stable Diffusion WebUI\footnote{\url{https://github.com/AUTOMATIC1111/stable-diffusion-webui}} to enrich its abilities to address the issues of the Chinese language. For application developers, our elastic inference toolkit~\emph{diffusers-api} makes it easy to deploy these models as RESTful web services. 
    It further supports our compiler performance optimization tool named~\emph{PAI-Blade}~\cite{DBLP:conf/eurosys/ZhuZZGZBYLDL21}. When the functionality is enabled, the image generation speed is improved by 2-3 times compared to native PyTorch implementation, while maintaining the effectiveness. 
\end{itemize}

To realize our promise for the~\emph{openness} of our research, we have taken the following actions to contribute \emph{PAI-Diffusion} to the community:
\begin{itemize}
\item All the diffusion models, LoRAs and ControlNets of~\emph{PAI-Diffusion}, including domain-specific ones, have been released in our organizational Hugging Face repository\footnote{\url{https://huggingface.co/alibaba-pai}}. 
Users can easily download and fine-tune all the models in order to support their own domain-specific applications.

\item Our Chinese WebUI and \emph{diffusers-api} toolkits have also been made publicly available\footnote{The Chinese WebUI extension is released under the EasyNLP~\cite{DBLP:conf/cikm/0001Q022,DBLP:conf/emnlp/WangQZLLWWHL22} framework:~\url{https://github.com/alibaba/EasyNLP/tree/master/diffusion/chinese_sd_webui}\\
The \emph{diffusers-api} repository:~\url{https://github.com/alibaba/diffusers-api}}, so that~\emph{PAI-Diffusion} models are easy to be deployed in other environments beyond the Alibaba Cloud ecosystem, for both designers and application developers.

\item All our models and codes are released under the Apache License (Version 2.0) to support both academic and commercial use.
\end{itemize}

\begin{figure*}
\begin{center}
    
  \includegraphics[width=.975\linewidth]{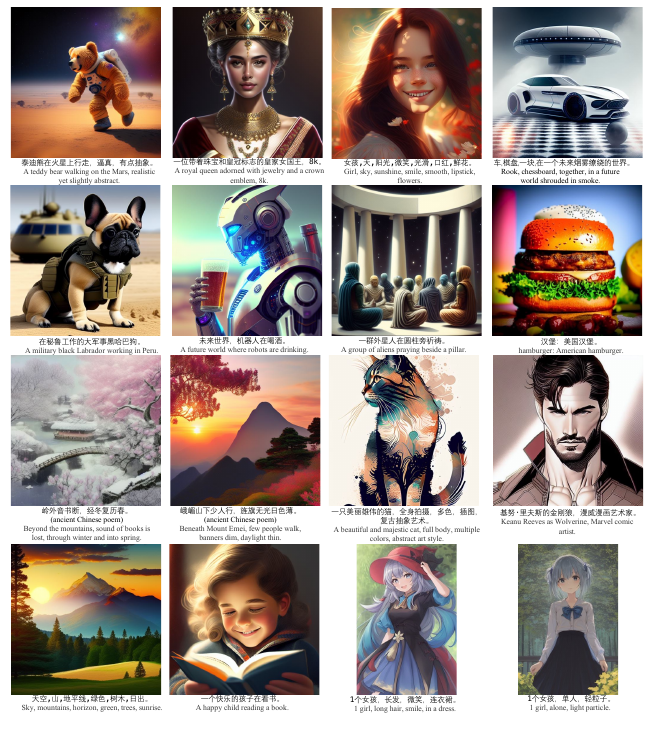}
    \vspace{-.5em}
  \caption{Some examples of the generated images by~\emph{PAI-Diffusion} models. Prompts are originally in Chinese and have been manually translated into English for reference.}
  \vspace{-1em}
  \label{fig:cases}
  \end{center}
\end{figure*}

\begin{table*}
\centering
\begin{small}
\begin{tabular}{llll} 
\hline
\bf Model Name  & \bf \#Parameters & \bf Image Size (Default) & \bf Domain\\ 
\hline
pai-diffusion-general-large-zh & 1.04B & 512$\times$512 & General purpose\\
\quad\quad\quad -controlnet-canny & 361M & 512$\times$512 & General purpose\\
\quad\quad\quad -controlnet-depth & 361M & 512$\times$512 & General purpose\\
pai-diffusion-general-xlarge-zh& 1.04B & 768$\times$768 & General purpose\\
\hline
pai-diffusion-artist-large-zh & 1.04B & 512$\times$512 & Artistic pictures\\
\quad\quad\quad -controlnet-canny & 361M & 512$\times$512 & Artistic pictures\\
\quad\quad\quad -controlnet-depth & 361M & 512$\times$512 & Artistic pictures\\
\quad\quad\quad -lora-poem & 25.5M & 512$\times$512 & Paintings for Chinese poems\\
\quad\quad\quad -lora-2.5d & 25.5M & 512$\times$512 & 2.5D-style arts\\
pai-diffusion-artist-xlarge-zh & 1.04B & 768$\times$768 & Artistic pictures\\
\hline
pai-diffusion-food-large-zh & 1.04B & 512$\times$512 & Chinese cuisines \\
pai-diffusion-anime-large-zh & 1.04B & 768$\times$512 & Cartoon characters (anime)\\
\hline
\end{tabular}
\end{small}
\caption{A summary of Chinese diffusion models, LoRAs and ControlNets released by us. 
}
\label{tab:model}
\end{table*}

\section{Models}

We introduce the models of~\emph{PAI-Diffusion}, a family of open Chinese diffusion models designed to address the challenges of generating high-quality images from Chinese textual descriptions.

\subsection{Model Zoo}

The model zoo of~\emph{PAI-Diffusion} consists of a collection of over ten open-source models, including base diffusion models, together with their corresponding LoRAs and ControlNets. A summary of these models is presented in Table~\ref{tab:model}, with a few cases of generated images presented in Figure~\ref{fig:cases}. Readers can also find these models from our organizational Hugging Face repository described previously.
Note that the list of models is not static. More new models will be added to the repository when they are ready to be released.

\subsection{Model Architecture}

To ensure full compatibility with the open-source community, our model architectures generally follow the~\emph{de facto} standard practice of Stable Diffusion~\cite{DBLP:conf/cvpr/RombachBLEO22} where a CLIP-based text encoder, a U-Net and a VAE model are leveraged to obtain text embeddings, generate image embeddings in the latent diffusion space and decode the image for output, respectively. Particularly, the architectures of the U-Nets and VAEs of our large and xlarge diffusion models are in line with Stable Diffusion 1.5 and 2.1, respectively.
Note that the difference between ``large'' and ``xlarge'' models is the size of generated images, rather than the number of parameters.

As for the Chinese language, we follow our previous work~\cite{DBLP:conf/acl/LiuLD0WZJJC023} to employ both 100 million text-image pairs from Wukong~\cite{DBLP:journals/corr/abs-2202-06767} and the largest Chinese KG available to us, i.e., OpenKG\footnote{\url{http://openkg.cn/}} as our knowledge source to pre-train a Chinese knowledge-enhanced CLIP model that better understand the morphologically rich semantics of the Chinese language. The resources for training general-purpose models are also the same as in ~\cite{DBLP:conf/acl/LiuLD0WZJJC023}. For more details, we refer our readers to the original paper. 

We have also collected a variety of domain-specific datasets to produce domain-specific diffusion models or LoRAs, depending on the volume of the corresponding datasets. Such domains include artistic pictures, paintings for Chinese poems, 2.5D-style arts, Chinese cuisines and cartoon characters. Note that we try our best to ensure that our models are built on legally and ethically sourced data.
We specifically filter out any images that may have the probability to exhibit some degree of ethical bias. We obtain the permission to use in-house datasets (such as Chinese cuisine) to train and release the corresponding models.

ControlNet~\cite{DBLP:journals/corr/abs-2302-05543} operates by incorporating a set of control vectors that encode specific image attributes or features. These control vectors are then incorporated into the residual blocks of the diffusion models. 
Furthermore, ControlNet allows for interactive image editing, enabling users to iteratively refine and modify the generated images, based on our WebUI toolkit.
To enable fine-grained control,~\emph{PAI-Diffusion} integrates several ControlNets with Chinese diffusion models, facilitating the modification and customization of generated images based on user preferences.
Currently, we have released the ControlNets based on canny edge detection~\cite{DBLP:journals/pami/Canny86a} and Midas depth maps~\cite{DBLP:journals/pami/RanftlLHSK22}. Users can train their own ControlNets, using the same methods as ControlNet for Stable Diffusion.

\subsection{Applications}

\emph{PAI-Diffusion} opens up a wide range of applications for the Chinese language. Here, we highlight some potential applications of~\emph{PAI-Diffusion}.

\subsubsection{Artistic Creations and Design}

\begin{figure}[h]
\centering
\includegraphics[width=0.5\textwidth]{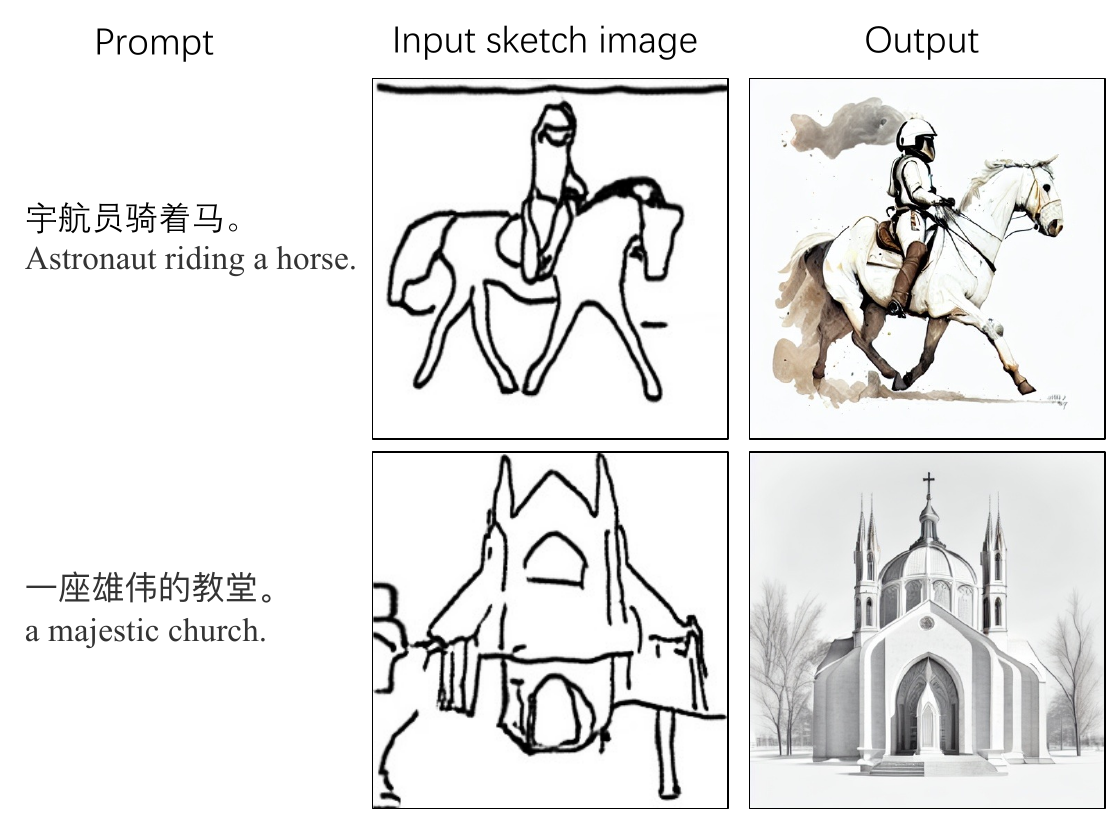}
\caption{Two examples of artistic creations from sketch images using the image-to-image pipeline.}
\label{fig:application_1}
\end{figure}

Diffusion models revolutionize the way sketch images are transformed into captivating artworks. With our models, users can witness their sketch images come to life based on the image-to-image pipeline. Examples can be found in Figure~\ref{fig:application_1}. We can see that our models empower artists to explore their creative boundaries and produce truly unique and mesmerizing artistic creations.

\subsubsection{Cultural Preservation and Heritage}

\begin{figure}[h]
\centering
\includegraphics[width=0.5\textwidth]{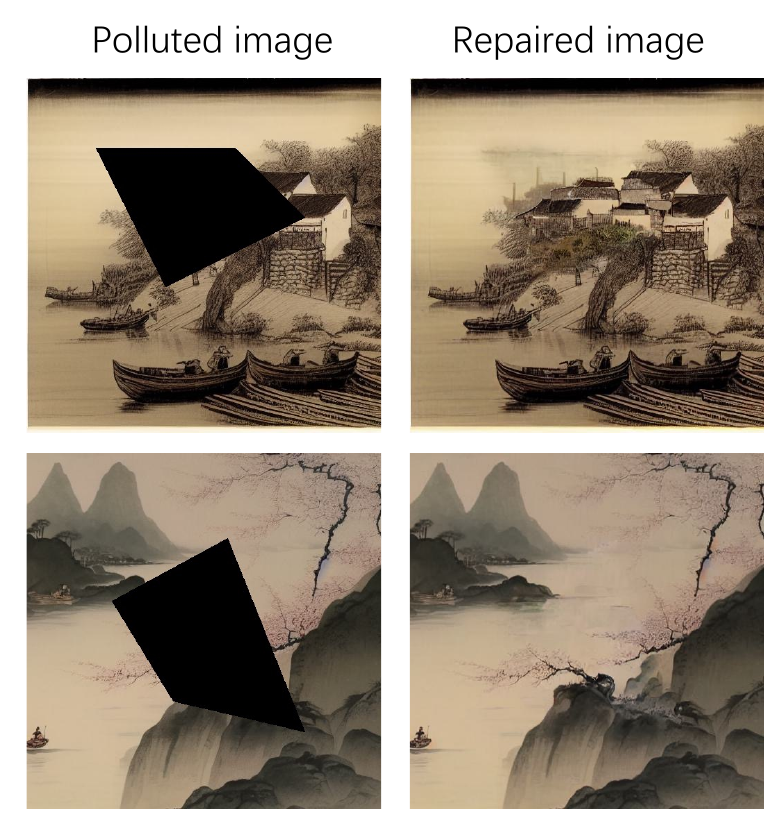}
\caption{Two examples of the restoration of ancient Chinese paintings via image in-painting.}
\label{fig:application_2}
\end{figure}

Our models present an elegant approach to restoring ancient Chinese paintings through image in-painting techniques. By harnessing the power of diffusion, these applications enable the recreation of missing or damaged areas in the paintings, seamlessly blending them with the original artwork (with examples shown in Figure~\ref{fig:application_2}). Through the application, ancient Chinese paintings once damaged or fragmented can be revitalized, allowing future generations to appreciate the cultural significance of invaluable artistic treasures.

\subsubsection{Visual Reality Generation}

\begin{figure}[h]
\centering
\includegraphics[width=0.485\textwidth]{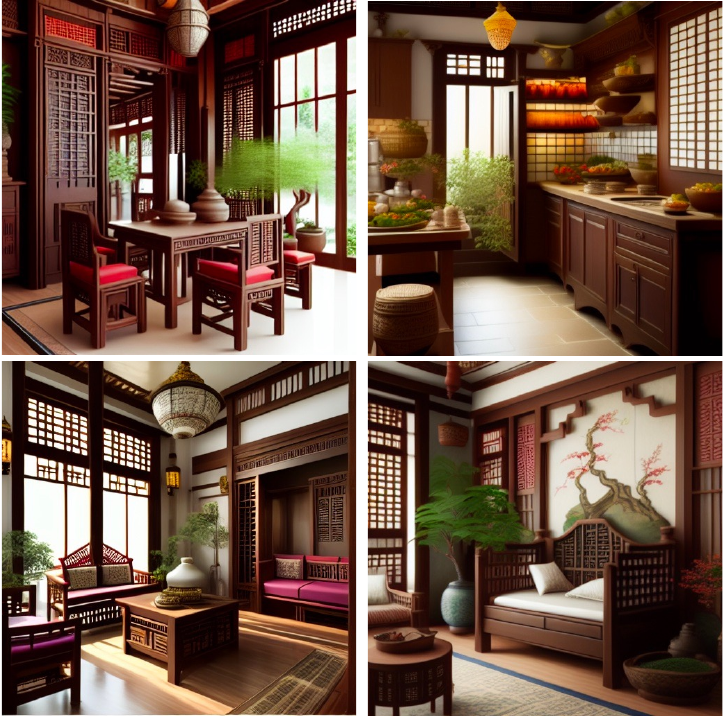}
\caption{Four examples of scene generation for Chinese garden architecture using the text-to-image pipeline.}
\label{fig:application_3}
\end{figure}

Diffusion models offer a cutting-edge approach to creating immersive virtual reality experiences. Take traditional Chinese garden architecture as an example (see Figure~\ref{fig:application_3}). Our model blends various elements such as architecture, house and furniture to generate virtual scenes reminiscent of ancient Chinese gardens. Such applications serve as a valuable tool for people to experience and appreciate the beauty of ancient Chinese culture regardless of their physical location.

\subsubsection{Others}

Apart from the three examples, our models can be harnessed in other domains such as fashion design, interior decor, and even product development. For instance, our model can create harmonious and visually stunning environments for interior decor.
In fashion design, our diffusion models can be applied to blend both traditional and contemporary styles, creating unique and captivating clothing designs.
We do not further elaborate.

\section{Toolkits}

\emph{PAI-Diffusion} provides user-friendly toolkits that facilitate the deployment and usage of diffusion models for Chinese text-to-image synthesis. In this section, we briefly introduce our two toolkits.

\subsection{Chinese WebUI Toolkit}

\begin{figure}
\centering
\includegraphics[width=0.5\textwidth]{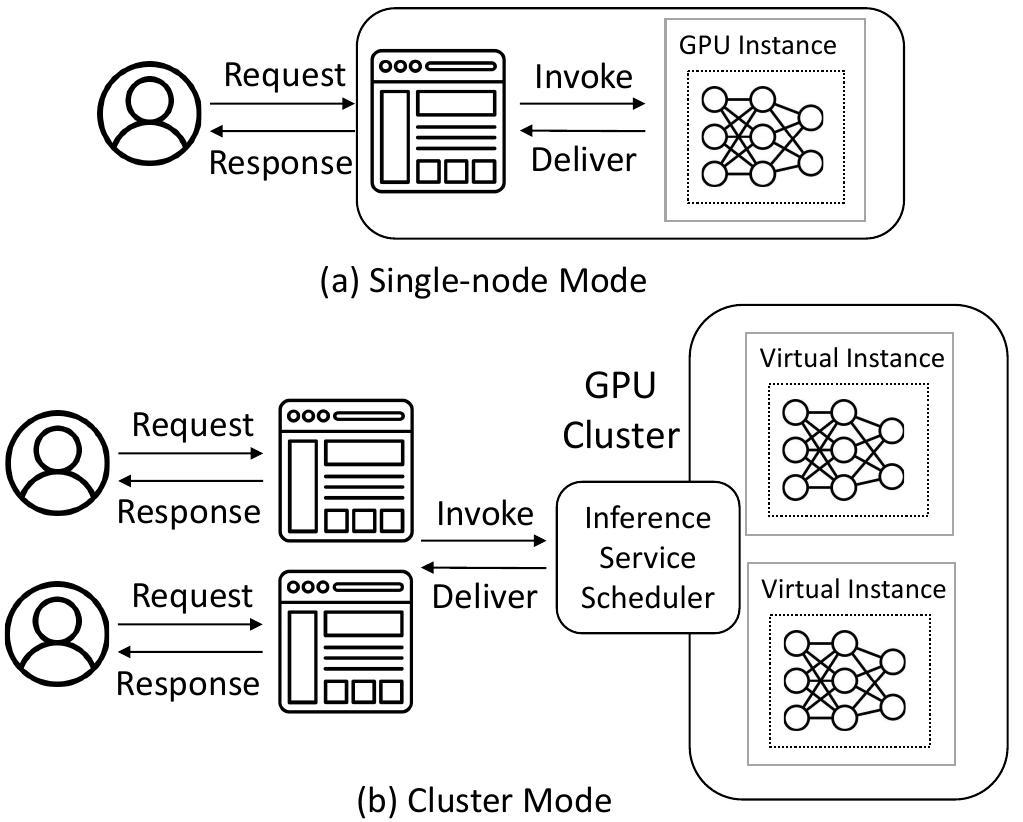}
\caption{The system architectures of two modes to deploy our Chinese WebUI toolkit online.}
\label{fig:modes}
\end{figure}

\begin{figure}
\centering
\includegraphics[width=0.455\textwidth]{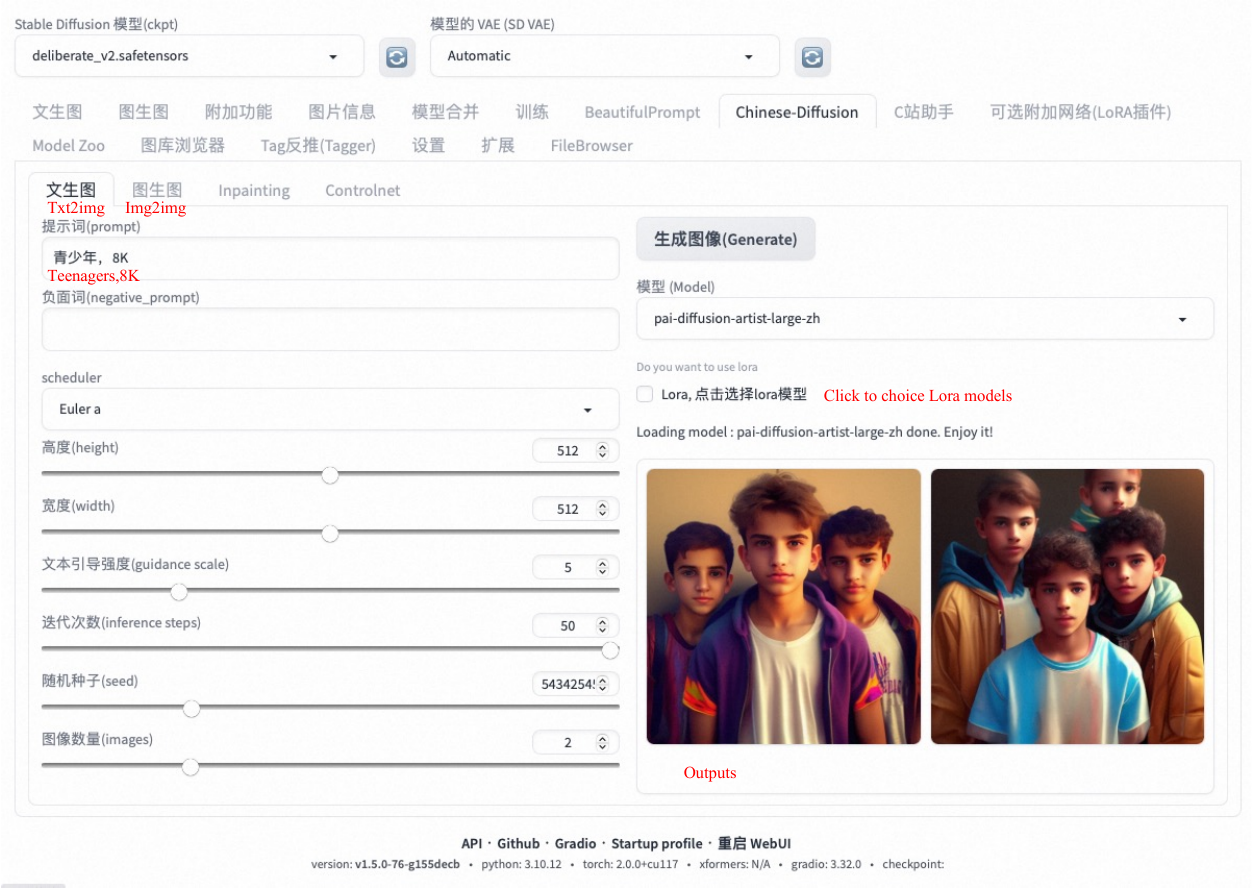}
\includegraphics[width=0.455\textwidth]{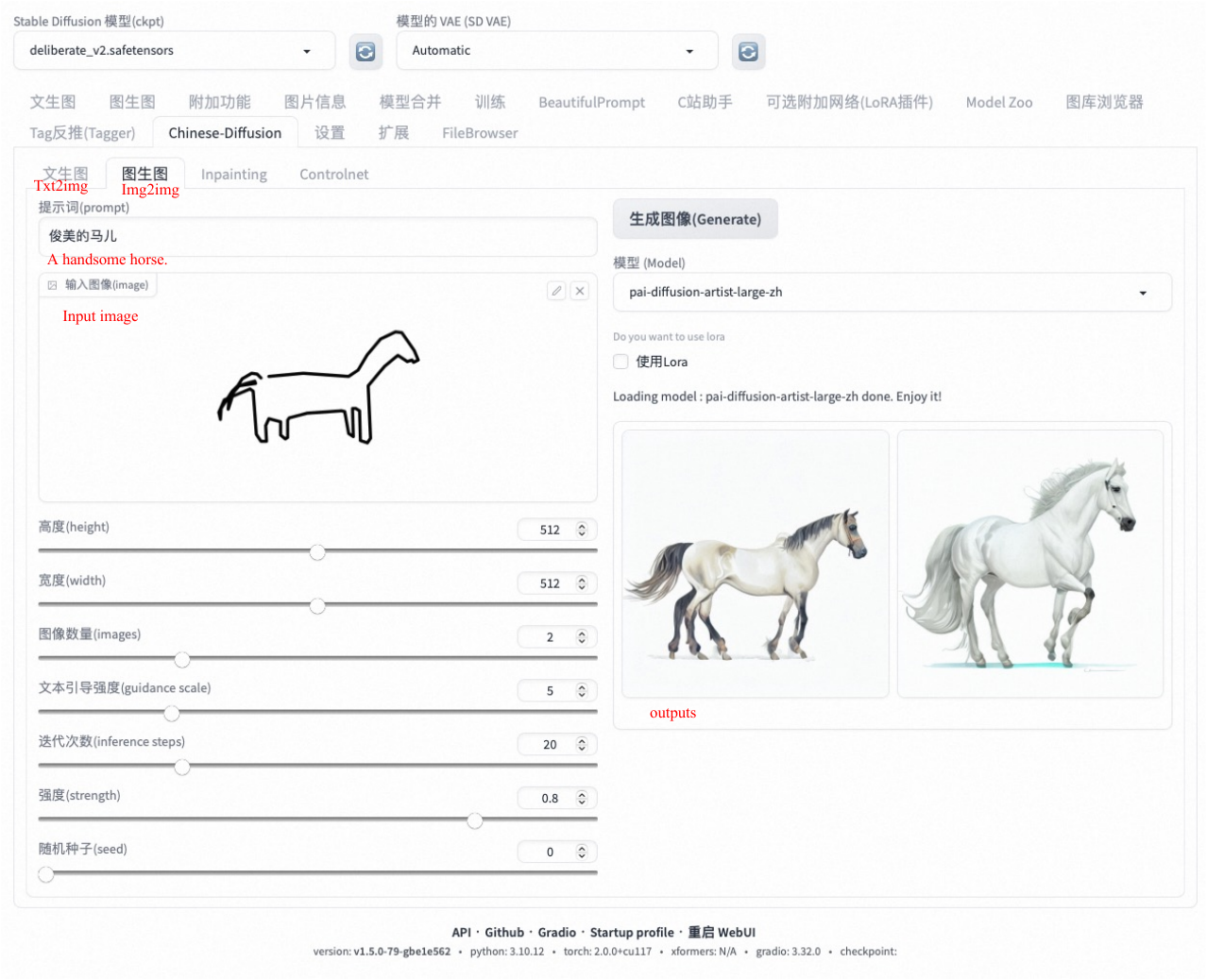}
\caption{The snapshots of Chinese WebUI for text-to-image and image-to-image synthesis.}
\label{fig:webui}
\end{figure}

The Chinese WebUI toolkit is an extension to the Stable Diffusion WebUI specifically developed to support our Chinese models. It offers a web-based graphical interface that allows users (especially designers without programming expertise) to interact with the diffusion models easily. The toolkit provides a seamless user experience, enabling users to perform image synthesizing and editing. 
Snapshots of the toolkit are shown in Figure~\ref{fig:webui}.
Additionally, our toolkit offers two modes:~\emph{single-node} mode and~\emph{cluster} mode (shown in Figure~\ref{fig:modes}), providing users with options based on their specific requirements and resources, introduced as follows:
\begin{itemize}
    \item \textbf{Single-node}: It is suitable for scenarios when users want to quickly set up and use the toolkit with exclusive computational resources. It is particularly useful for individual users or small-scale deployments.

    \item \textbf{Cluster}: Leveraging the elastic inference service of PAI, the cluster is geared towards users who require high scalability and performance. In this mode, the toolkit utilizes a cluster of nodes to handle the workload, enabling parallel processing and efficient utilization of resources. It ensures efficient scaling and responsiveness, making it suitable for enterprise-level deployments or applications that require extensive computational power.
\end{itemize}

\subsection{The~\emph{diffusers-api} Toolkit}

The~\emph{diffusers-api} toolkit is built upon the Hugging Face~\emph{diffusers} library\footnote{\url{https://github.com/huggingface/diffusers}} that provides a cloud service implementation for~\emph{PAI-Diffusion} models based on the PAI elastic inference service.\footnote{Note that~\emph{diffusers-api} is also compatible with original Stable Diffusion, which is not the focus of this paper.} Especially, we implement several inference pipelines customized for our Chinese models. 
Users can take advantage of a number of functions when sending HTTP requests, including text-to-image synthesis, image-to-image synthesis, image in-painting, image editing, etc. Below we show a sample request body for calling the~\emph{diffusers-api} service, with Chinese prompts manually translated into English for better understanding:

\begin{small}
\begin{verbatim}
{
 "task_id": "001",
 "prompt": "romantic starry sky",
 "negative_prompt": "noise, low-quality",
 "func_name" : "t2i",
 "steps": 25,
 "image_num": 1,
 "width": 512,
 "height": 512,
 "use_base64": True
}
\end{verbatim}
\end{small}
where ``\texttt{func\underline{ }name}'' refers to the specific functionality that the user wish to use (``\texttt{t2i}'' refers to text-to-image in this case).
We release the source code of~\emph{diffusers-api} to provide users with the ability to quickly develop customized API services. This allows for easier implementation of any desired functionality, tailored to specific requirements. For example, users can build and customize their own schedulers~\cite{DBLP:journals/corr/abs-2305-14677}.

In addition,~\emph{diffusers-api} leverages the AI compiler~\emph{PAI-Blade}~\cite{DBLP:conf/eurosys/ZhuZZGZBYLDL21} for inference optimization. It significantly reduces the end-to-end latency of the inference processes and the GPU memory consumption,  which ensures improved performance and efficiency in generating high-quality images, without any precision loss in computation. To verify the correctness of our argument, we have deployed our Chinese diffusion model (the large version) online using an NVIDIA A10 GPU with 50 sampling steps for inference. The generated image size is fixed to 512$\times$512. We repeat the experiments in 20 times and report the averaged results in Table~\ref{tab:sample}, which clearly prove the correctness of our claim.

\begin{table}
\centering
\begin{small}
\begin{tabular}{lll} 
\hline
\bf Settings  & \bf PyTorch Native & \bf\emph{diffusers-api}\\ 
\hline
Inference time (s) & 6.34 & \bf 2.96\\
GPU memory (GB) & 6.94 & \bf 5.56\\
\hline
\end{tabular}
\end{small}
\caption{The performance of~\emph{diffusers-api} for online deployment of our diffusion model.}
\label{tab:sample}
\end{table}

\section{Conclusion}

This paper presents~\emph{PAI-Diffusion} to address the challenges in Chinese text-to-image synthesis.~\emph{PAI-Diffusion} integrates both general and domain-specific Chinese diffusion models, LoRAs and ControlNets, enabling the generation of contextually relevant images. Moreover,~\emph{PAI-Diffusion} ensures seamless integration with our cloud platform, offering accessible and scalable solutions. The release of models further encourages collaboration and innovation in the field. The public availability of Chinese WebUI and~\emph{diffusers-api} toolkits simplifies deployment in various environments. Our advancements pave the way for further research and development in Chinese text-to-image synthesis and relevant applications.

\newpage

\section*{Acknowledgements}

The authors would like to thank other members of the Machine Learning Platform for AI (PAI) team of Alibaba Cloud for the help of this work.
This work was in part supported by the National Natural Science Foundation of China under grant number 62202170, Fundamental Research Funds for the Central Universities under grant number YBNLTS2023-014, Alibaba Group through Alibaba Innovative Research Program, and Alibaba Cloud through Research Talent Program with South China University of Technology.

\section*{Limitations}

There are still some limitations that should be acknowledged. The effectiveness of our models heavily relies on accurate mappings between textual descriptions and visual representations.
While efforts have been made to include a variety of domain-specific models, there may still be domains or contexts for which specialized models are not available. This limitation restricts the full potential in generating highly relevant and specific images for a wide range of applications. We suggest that users should further fine-tune our models if necessary.

\section*{Ethical Considerations}

It is important to consider the ethical implications associated with its use and deployment. Diffusion models like~\emph{PAI-Diffusion} learn from large datasets, which may inadvertently contain biases present in the data. It is crucial to be aware of and mitigate any biases that may be perpetuated in the generated images. In addition,~\emph{PAI-Diffusion} has the potential to be misused for unethical purposes, such as generating inappropriate or harmful contents. Therefore, users should be encouraged to adhere to ethical standards and abide by terms and regulations when utilizing~\emph{PAI-Diffusion} models.





\end{document}